\documentclass[preprint,12pt]{elsarticle}
\usepackage{xcolor}
\usepackage{makecell} 
\usepackage{diagbox}
\usepackage{adjustbox}  

\usepackage{amssymb}
\usepackage{amsmath}

\journal{Mechanism and Machine Theory}

\begin{document}

\begin{frontmatter}



\title{Model Analysis And Design Of Ellipse Based Segmented Varying Curved Foot For Biped Robot Walking}
\author[inst1]{Boyang Chen}
\author[inst1]{Xizhe Zang}
\author[inst1]{Chao Song\corref{cor1}}
\author[inst1]{Yue Zhang}
\author[inst1]{Jie Zhao}
\cortext[cor1]{Corresponding author. Email: songchao@stu.hit.edu.cn}
\affiliation[inst1]{organization={School of Mechatronics Engineering, Harbin Institute of Technology},
            addressline={92 West Dazhi Street}, 
            city={Harbin},
            postcode={150001}, 
            state={Heilongjiang},
            country={China}}

\begin{abstract}
This paper presents the modeling, design, and experimental validation of an Ellipse-based Segmented Varying Curvature (ESVC) foot for bipedal robots. Inspired by the segmented curvature rollover shape of human feet, the ESVC foot aims to enhance gait energy efficiency while maintaining analytical tractability for foot location based controller. First, we derive a complete analytical contact model for the ESVC foot by formulating spatial transformations of elliptical segments only using elementary functions. Then a nonlinear programming approach is engaged to determine optimal elliptical parameters of hind foot and fore foot based on a known mid-foot. An error compensation method is introduced to address approximation inaccuracies in rollover length calculation. The proposed ESVC foot is then integrated with a Hybrid Linear Inverted Pendulum model-based walking controller and validated through both simulation and physical experiments on the TT II biped robot. Experimental results across marking time, sagittal, and lateral walking tasks show that the ESVC foot consistently reduces energy consumption compared to line, and flat feet, with up to 18.52\% improvement in lateral walking. These findings demonstrate that the ESVC foot provides a practical and energy-efficient alternative for real-world bipedal locomotion. The proposed design methodology also lays a foundation for data-driven foot shape optimization in future research.
\end{abstract}

\begin{graphicalabstract}
\includegraphics[width=1.0\textwidth]{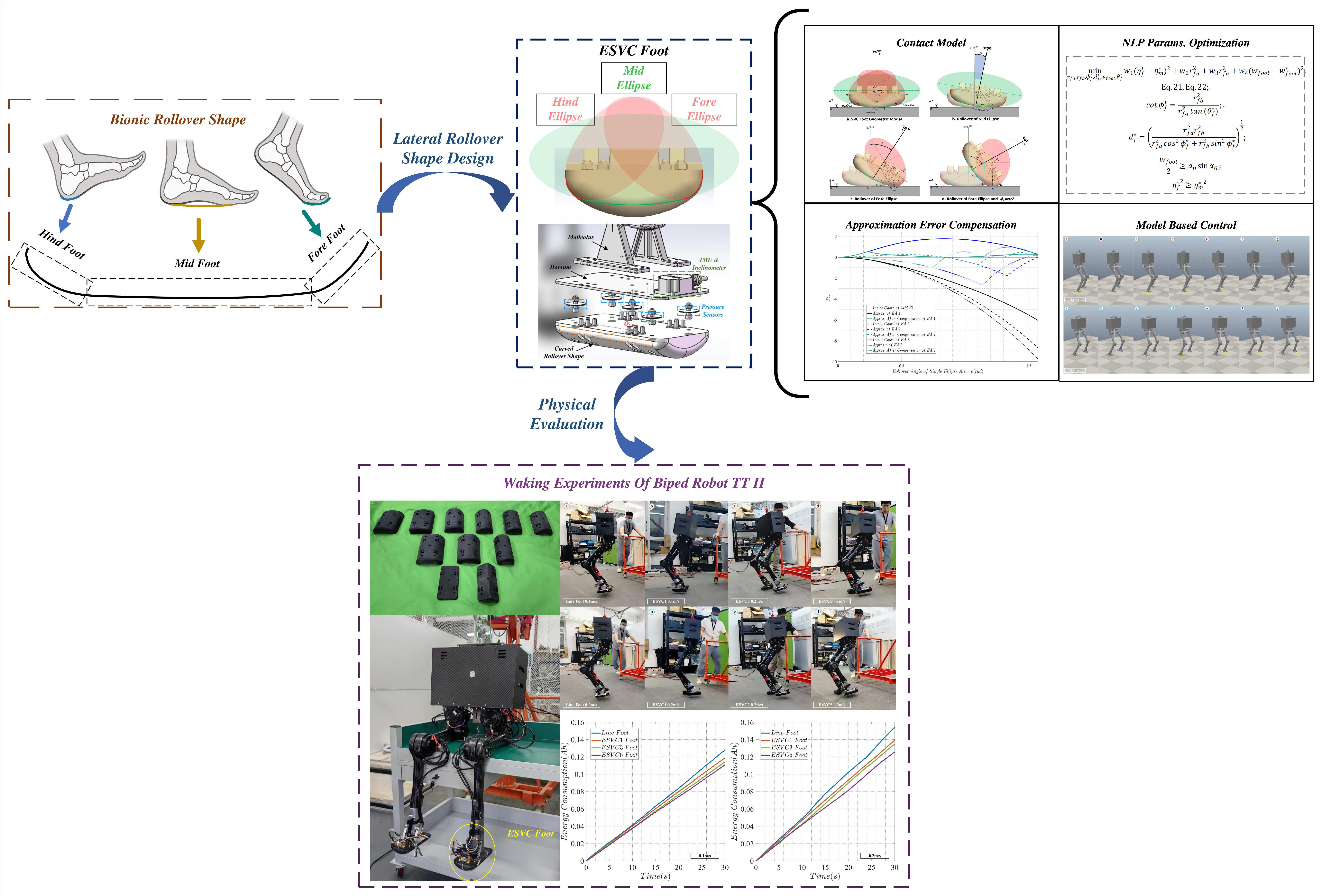}
\end{graphicalabstract}

\begin{highlights}
\item A ellipse-based varying curvature robotic foot and its analytical contact model, formulated only using elementary functions, are proposed. The approximation error of the contact model is systematically analyzed and compensated.
\item A nonlinear programming based design approach for ESVC Foot is proposed which resolves the parameter coupling problem inherent in single elliptical arc foot design.
\end{highlights}


\end{frontmatter}


\section{Introduction}
\label{sec1}
Legged robots, especially in the quadruped and biped fields, have made significant progress recently, with a number of impressive movements now achievable. Runtime has become one of the primary limitations hindering practical performance. The foot, as the only part of the body interacting with the external environment, not only participates in the evolution of the entire dynamic system, but also directly affects the gait energy efficiency, especially in control methods based on simplified models. Therefore, it is important to design a robot foot that achieves efficiency comparable to a biological foot\cite{reference1,reference2,reference3}.

To explore the effect of foot-ground contact on bipedal locomotion, researchers of human 
kinesiology and prosthetics have conducted a series of studies on biological foot structures,  center of pressure trajectory and rollover shape\cite{reference4,reference5,reference6,reference7,reference8} and they observed that both the human foot or prosthetic foot shape can be segmented into three parts: the fore-foot, mid-foot, and hind-foot, with each part having a varying curvature. Specifically, the fore-foot and mid-foot are more curved, while the mid foot is relatively flat. This segmented variable curvature morphology allows human walking to achieve higher energy efficiency. In the filed of legged-robotics, since researchers often aim for lower control complexity, simplified foot designs are preferred to reduce the complexity of foot-ground contact, such as point foot. With recent advancements in the field of legged robotics, such as the impressive performance of robotic dogs and numerous humanoid robots, we think that research on robot foot will return to the same level of importance and attention, comparable to that of joint design and dexterous hands\cite{reference9, reference10}.

From the perspective of bipedal robots' foot morphology, the current mainstream foot shapes can be categorized as arched foot\cite{reference11}, point foot\cite{reference12}, flat foot\cite{reference13}, line foot\cite{reference14}, arc/curved foot\cite{reference7,reference9}, biomimetic foot\cite{reference15} and flexible heterogeneous foot\cite{reference10}. Point foot is the simplest, as its mass can be ignored relative to the overall robot and the contact point can be considered fixed with ground which makes the kinematic and dynamic models easy to analyze and represent\cite{reference16,reference17}. Hence, it is beneficial for model-based control methods such as Inverse Pendulum (IP). However, whether from the sagittal plane or the coronal plane, there is only a single contact point with the ground, which not only makes it impossible to maintain static balance but also fails to provide a frictional force/torque, especially when upper body manipulation is involved. In other words, even if there is no positional change in joints, the robot's posture may still tilt. Moreover, according to the research\cite{reference7}, the energy efficiency of point foot is often lower than that of flatter foot types. In contrast, the more widely used flat foot, especially preferred by humanoid robots, maintains a flat line contact with ground in both the saggital and coronal planes to provide sufficient contact friction force while performing upper body operations, such as Atlas\cite{reference18} and Digit\cite{reference19}. Using overly flat foot shape during walking introduces greater instantaneous impact forces which will make the gait less natural. Additionally, flat foot typically requires more active degrees of freedom at ankle, which increase the joint weight and weaken mass centralization. Although some robots transfer the foot drive to the legs or waist via external linkages, such as Digit\cite{reference19} and KUAVO\cite{reference20}, the additional transmission mechanism requires the robot to occupy more space during walking. Another widely used foot shape is the line foot, which maintains the advantages of light weight and simple structures similar to the point foot and is able to provide yawing friction for preventing self-rotation due to line contact with the ground in sagittal plane\cite{reference14,reference15,reference21}. But it is still considered a point foot in coronal plane which energy efficiency remains insufficient. To improve gait energy efficiency, arc/curved shaped foot is often considered and the foot-ground contact point will rollover along the curved shape during support phase. Since it extends the contact duration when impact, not only is the normal pressure smaller, but it also reduces energy dissipation at the collision. The energy efficiency of arc/curved shaped feet is superior to that of point feet, with flat foot falling in between, and arc-shaped feet have also been shown to enable a more human-like walking gait\cite{reference7,reference11,reference22,reference23,reference24}. It's worth noting that the curved foot with constant curvature are not only unrealistic in biological foot, but their energy efficiency also differs significantly from that natural foot shapes\cite{reference6,reference9,reference15}. Moreover, when extending the constant curvature foot to three dimensions, it will resemble those of point feet. To pursue higher energy efficiency, researchers have incorporated the segmented variable curvature(ESVC) feature from biological foot shape into robot foot design and similar to biological counterpart, robot foot is also divided into three parts: mid-foot, fore-foot, and hind-foot\cite{reference7,reference9,reference24,reference25,reference26,reference27}. Through the observation of the three parts of biological foot, it was found that mid-foot is the flattest, while the forefoot and hind-foot are more curved. In order to approximate the energy efficiency of a biological foot, this curvature characteristic is preserved in the design of the ESVC robot foot. Mahmoodi et al.\cite{reference6} developed a ESVC foot shape based on piecewise polynomial and confirmed that the segment curvature of the foot shape has a significant impact on gait period, walking speed, and inter-leg angles through bifurcation diagrams. Besides polynomial profiles, ellipse is also used to design rollover shapes due to its varying curvature feature. Silva et al.\cite{reference25} utilized two circle arc and one flatten profile to form the rollover shape. Integrated with a 3D Linear Inverted Pendulum Model controller, the robot could reducing energy expenditure by almost 25\% in simulation. A completely flatten mid-foot neither meets the energy efficiency requirements nor the bionic design. Smyrli et al.\cite{reference7} proposed a ESVC foot using the elliptical natural varying curvature characteristic and, through numerical methods, investigated the relationship between the elliptical eccentricity and gait efficiency, stability, impact force and walking speed by traversing the length. They found that for a targeted forward gait velocity, less pointy foot shapes, associated with larger eccentricity for a certain minor axis, allow for larger gait energy efficiency. However, the analytical form of the kinematic model for the contact points was not provided due to the incomplete elliptic integral and the single-ellipse design can not allow the rollover shape been modified arbitrarily for application or bionic design. In \cite{reference27} and \cite{reference28}, researchers attempted to use a segmented convex rollover shape to simulate biological feet and derive dynamic models. However, finding a suitable foot shape for robots is challenging, and thus, analytically expressing the contact model remains difficult. For the arched foot, biomimetic foot, and flexible heterogeneous foot, which are designed for specific scenarios, are not within the discussion of this paper due to the complexity of their design and implementation, as well as the difficulty in analyzing their effect on bipedal locomotion.

Combining the principle of segmented bionic design and the varying curvature characteristics of ellipse, an Ellipses based Segmented Variable Curvature Foot(ESVC Foot) is proposed. Both the foot-ground contact model, design methodology, and accuracy of ESVC Foot have been discussed. The main contributions of this paper are as follows: Firstly we investigate the kinematic model of contact point rollovering along a single ellipse arc and the space transformation matrices of ESVC foot based on elementary functions are given analytically. Secondly, a nonlinear program is used to design the ESVC foot, especially in determining the fore/hind foot. Then we discuss the error characteristics of contact model and provide a compensation scheme. Finally, we integrated the ESVC Foot with a model-based bipedal motion control framework and conduct real robot experiments to confirm the improvement in energy efficiency.

The organization of this paper is as follows: In Sec II.a, we first clarify the motivation, assumptions, and definitions of the key coordinate systems of the ESVC Foot. In Sec II.b, we derive contact model for single elliptical arc. The spatial transformation matrices between key coordinate systems is also provided. Then we demonstrate the nonlinear program for ESVC Foot design and present the entire analytical contact model of ESVC foot (Sec II.c). In Sec II.d, error characteristic and compensation is investigated. The integration of the ESVC Foot and model-based control framework(HLIP) is formulated in Sec III. In Sec IV, we conduct directional walking experiments on the TT II (Test Traveller II) robot with 5 kind of feet, The conclusion and future work are presented in Sec V.
\begin{figure}[ht]
	\centering
	\includegraphics[width=1.0\textwidth]{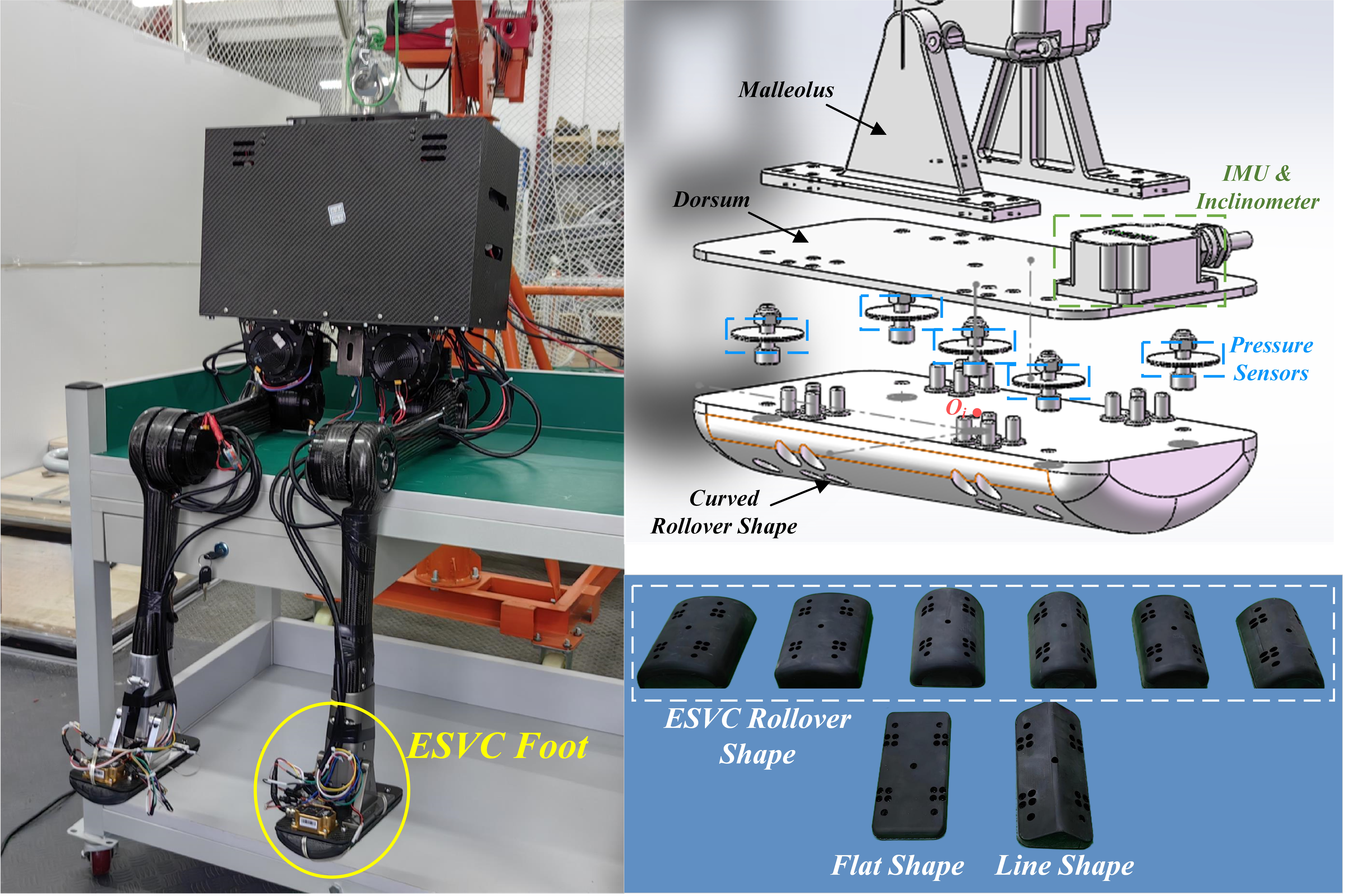}
	\caption{a.Biped Robot TT II With ESVC Foot; b.The Assembly Drawing Of ESVC Foot; c.Rollover Shape Of Different Foot Types}\label{fig_TTII_ESVC_total}
\end{figure}

\section{Geometric Analysis and Contact Model of ESVC Foot}
\label{sec2}
\subsection{Preliminary}
\label{subsec2.1}
\subsubsection{Motivations and Question Define}
\label{subsec2.1.1}
Researches on human foot morphology has revealed that biological foot can be composed of fore-foot, mid-foot and hind-foot and the feature of segmented varying curvature in the rollover shape is a key factor for achieving efficient gait. Compared to other rollover shape fitting profiles, the elliptical arc is considered one of the ideal curves for fitting biological feet due to its inherent varying curvature feature. However, as described in \cite{reference26,reference27}, a single-segment elliptical arc is insufficient to mimic the morphological characteristics of biological counterpart. Therefore, the traditional elliptical arc foot has limited effectiveness in improving gait energy efficiency. In addition, the design and modeling analysis of the elliptical foot remain challenging. In other words, due to the presence of elliptical integrals, analytically expressing the contact model of the elliptical arc foot is still difficult. Therefore, we propose an ellipse-based segmented varying curvature foot(ESVC foot) and discuss its model analysis and design methods.
\subsubsection{Foot Shape Specification}
\label{subsec2.1.2}
Biomimetic foot and prosthetic foot, based on biological data as a template, are often too complex for robot design. Furthermore, since those rollover shapes are adapted to natural biomechanics, directly copying biological data to design robot foot is not appropriate. Nevertheless, there is a consensus that the curvature is varying for each part and hind/fore foot is more curved than mid foot which is naturally resembles an elliptical arc. 
Considering practical applications, ESVC foot is designed like a flat-bottom boat,as shown in Figure\ref{fig_TTII_ESVC_total}.In the sagittal plane, the ESVC foot executes a line contact similar to a flat foot,providing sufficient friction during upper limb operations.The contact point is located at the mid point of the contact line. In the coronal plane, three elliptical arcs are used to fit the segmented varying curvature characteristics of the biological foot,namely hind/mid/fore ellipses for hind/mid/fore foot, respectively. The eccentricity of mid ellipse is larger than hind/fore ellipse, and hind ellipse and fore ellipse are symmetric with respect to the mid line.
\subsubsection{Assumptions \& Key Frames Define}
\label{subsec2.1.3}
Before analyzing the geometric and kinematic of ESVC foot, some assumptions need to be introduced:

\textit{F1.The contact of foot-ground is assumed to be rigid and the foot will not deform};

\textit{F2.The foot roll on the ground without slipping during the support phase}.
\\When the assumptions are satisfied, the contact point will rollover along the sole surface.Due to the varying curvature,the state of the center of mass (COM) and the swing foot is time variant. Hence, the contact model is also time variant.

To analyze this time-varying model, we divide the entire robotic system into internal space and external space by the foot frame which is fixed at the center of foot upper surface.In the internal space, the state of the robot's center of mass and the swing foot depends only on the joint positions and velocities, making it easy to calculate. In external space, it is important to determine the transition between the foot frame and contact point. In other words, once the transformation matrix can be expressed analytically, the COM state and the pre-contact point of the swing foot relative to the contact point frame can be calculated. Moreover, for more complex tasks, such as climbing stairs, precise foot placement planning is essential. Therefore, the transition between the world frame and the contact frame, specifically the rollover length, must be determined.For simplicity, we use a contact frame with the foot roll angle of 0 degrees to replace the world frame. At this point, the three key frames of the ESVC foot can be defined as frame $O_i$, frame $O_c$, and frame $C$, as shown in Figure\ref{fig2}.

\subsection{Geometric Analysis of Single Elliptic Arc}
\label{subsec2.2}

\begin{figure}[h]
	\centering
	\includegraphics[width=1.0\textwidth]{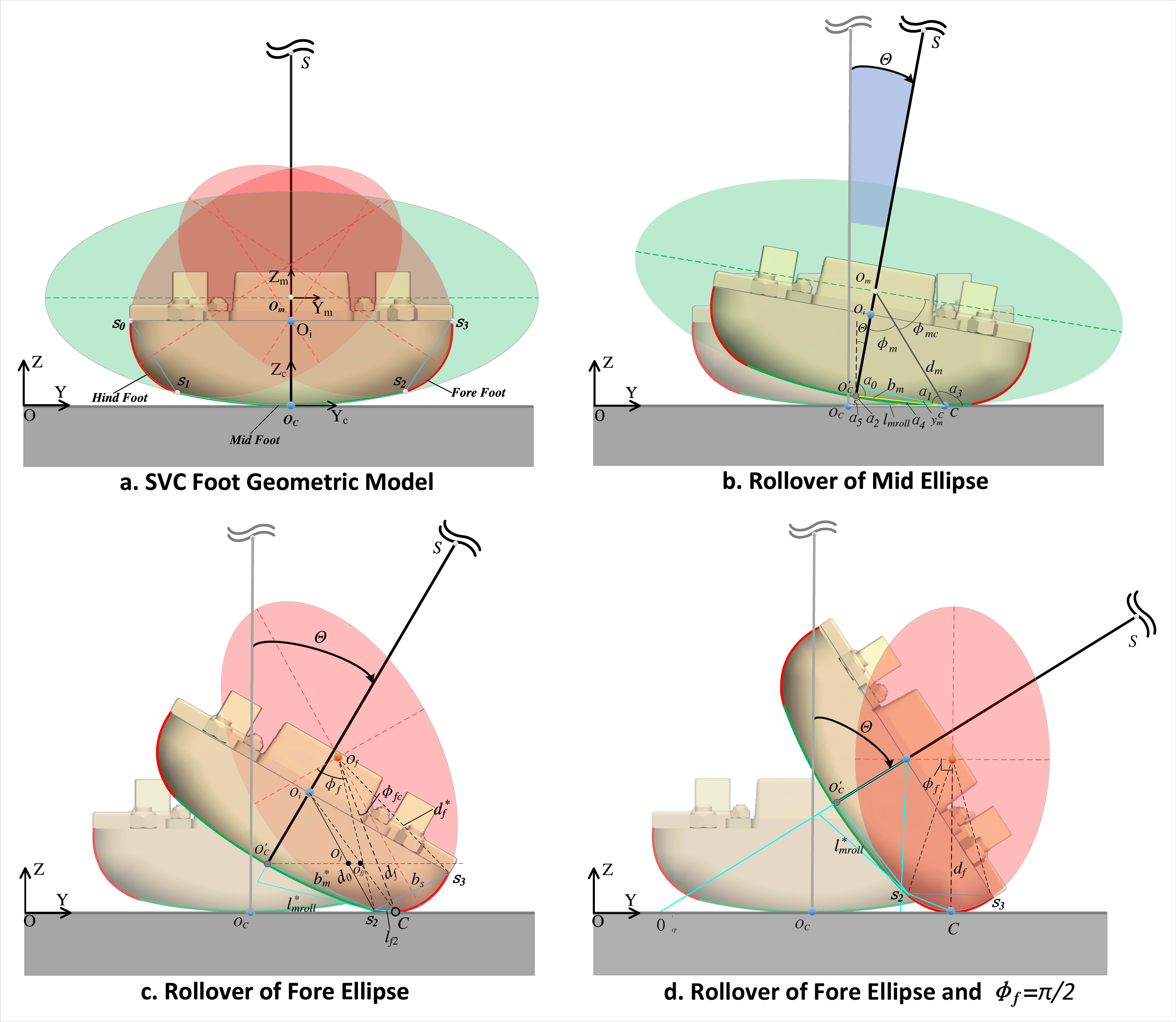}
	\caption{Rollover Process of ESVC Foot}
	\label{fig2}
\end{figure}

For the elliptical-based segmented varying curvature foot proposed in this paper, each segment consists of an elliptical arc, where the hind-foot and fore-foot are symmetric. The overall rollover process of the ESVC foot is described in Figure\ref{fig2}. Fig2.a shows the case when the foot rollover angle is 0 rad, i.e., when the foot is vertically contact the ground at $O_c$. Fig2.b depicts the scenario when the contact point rolls within the mid-foot. Fig2.c illustrates the case when the contact point rolls in the forefoot or hind-foot. Fig2.d represents the case just before the foot lifts off the ground which can be considered the toe is to leave the ground.Then the rollover shapes are described by the mid ellipse,hind ellipse and fore ellipse, respectively.

For any single-segment elliptical arc, the transformation matrix between the three key frames is related to the rollover length and the rollover angle of contact point. The rollover angle is defined as the angle between the line connecting the contact point and the foot frame, and the horizontal axis. Using the mid ellipse as an example we derive the transformation matrix of key frames.

Consider the case when the ESVC foot tilts from left to right, with the rollover range within the mid ellipse, as shown in Fig2.b. In this case, the minor axis of the ellipse coincides with the vertical axis of the foot surface. $\theta$ represents roll angle of the foot which can be obtained from inclinometer. The roll angle of mid ellipse is expressed as $\theta_m=|\theta|$. Then the rollover angle of mid ellipse $\phi_m$ and its complementary angle $\phi_{mc}$ can be determined as:
\begin{equation}
	\label{eq.1}
	\tan\phi_{mc} = \frac{r_{mb}^2}{r_{ma}^2\tan\theta_m}
\end{equation}
\begin{equation}
	\label{eq.2}
	\phi_{m} = \frac{\pi}{2}-\phi_{mc}
\end{equation}
The distance from center ellipse $O_m$ to the contact point $C$ can be expressed as:
\begin{equation}
	\label{eq.3}
	d_{m} = (\frac{r_{ma}^2r_{mb}^2}{r_{ma}^2\cos^2\phi_m+r_{mb}^2\sin^2\phi_m})^\frac{1}{2}
\end{equation}
The Y-coordinate of the contact point $C$ relative to the mid ellipse center frame $O_m$, as shown by the cyan dashed line in Fig2.b, can be calculated as:
\begin{equation}
	y_{C}^m = d_{m}\sin\phi_m
\end{equation}
The elliptical parameter angle satisfies:
\begin{equation}
	\lambda_m = \sin^{-1}\frac{y_{C}^m}{r_{ma}}
\end{equation}
The rollover length of contact point $C$, namely the ellipse arc length of $\phi_m$ is define through the incomplete elliptic integral:
\begin{equation}
	l_{mroll} = r_{ma}\int_{0}^{\lambda}\sqrt{1-e_m^2\sin^2\lambda_m}d\lambda_m
\end{equation}
Unfortunately, this integral cannot be calculated directly, a approximation of the rollover length using elementary functions as described in Eq.\ref{eq.7} derived from \cite{reference29}.
\begin{equation}
	\label{eq.7}
	l_{mrolla} = r_{ma}(\lambda_m-(\lambda_m-\sin\lambda_m)\frac{2k_m}{\pi}-\xi(\pi-(\pi-2)\frac{2k_m}{\pi}-2E_m)\frac{\lambda_m-\lambda_m^*}{\pi-2\lambda_m^*})
\end{equation}
Let $e_m$ be the eccentricity of the mid ellipse, then the modulus angle $k_m=\sin^{-1}e_m$. $E_m$ is a parameter depend on$k_m$, and $\xi$ is binary variable of threshold of parameter angle $\lambda_m^*$:
\begin{equation}
	E_m = \frac{{\pi}r_{mb}+4(r_{ma}-r_{mb})}{4r_{ma}}\left(1+(\frac{r_{mb}}{r_{ma}})^{\frac{3}{2}}\right)
\end{equation}
\begin{equation}
	\lambda_m^* = \frac{36k_m+13\pi}{52}
\end{equation}
\begin{equation}
	\label{eq.10}
	\xi = 
	\begin{cases}
		0, & \lambda_m < \lambda_m^* \\
		1, & \lambda_m \ge \lambda_m^*
	\end{cases}
\end{equation}
Since both the frame $O_C$ and the contact point frame $C$ are aligned to the world frame, the transformation matrix between the two frames can be derived analytically using:
\begin{equation}
	{}_{O_C}^{C}\!T_m = 
	\begin{bmatrix}
		1 & 0 & 0 & 0\\
		0 & 1 & 0 & sgn(\theta)l{mrolla}\\
		0 & 0 & 1 & 0\\
		0 & 0 & 0 & 1\\
	\end{bmatrix}
\end{equation}
For the transition of frame $C$ and frame $O_i$, We solve it using the frame $O_C^{\prime}$. The inside chord length $b_m$, as shown by the yellow dashed line in Fig2.b, is determined by:
\begin{equation}
	b_m = (d_m^2 + r_{mb}^2 - 2d_mr_{mb}\cos\phi_m)^{\frac{1}{2}}
\end{equation}
Then the auxiliary angle $\alpha_0\sim\alpha_5$ is:
\begin{equation}
\left\{
\begin{aligned}
	\alpha_0 &= \sin^{-1} \left( \frac{d_m \sin \phi_m}{b_m} \right) \\
	\alpha_1 &= \sin^{-1} \left( \frac{r_{mb} \sin \phi_m}{b_m} \right) \\
	\alpha_2 &= \pi - \alpha_0 \\
	\alpha_3 &= \pi - \phi_{mc} - \theta_m \\
	\alpha_4 &= \pi - \alpha_3 - \alpha_1 \\
	\alpha_5 &= \pi - \alpha_2 - \alpha_4
\end{aligned}
\right.
\end{equation}
The Y-coordinate and Z-coordinate the origin of frame $O_C^{\prime}$ in frame $C$ can be represented: 
\begin{equation}
	\left\{
	\begin{aligned}
		y_{O_C^{\prime}}^C &= sgn(\theta)b_m\cos\alpha_4 \\
		z_{O_C^{\prime}}^C &= b_msin\alpha_4 \\
	\end{aligned}
	\right.
\end{equation}
Hence, the transformation matrix of frame $O_C^{\prime}$ to frame $C$ is:
\begin{equation}
	{}_{O_C^{\prime}}^{C}\!T_m = 
	\begin{bmatrix}
		1 & 0 & 0 & 0\\
		0 & \cos\theta_m & \sin\theta_m & y_{O_C^{\prime}}^C\\
		0 & -\sin\theta_m & \cos\theta_m & z_{O_C^{\prime}}^C\\
		0 & 0 & 0 & 1\\
	\end{bmatrix}
\end{equation}
As the transformation matrix of frame $O_i$ and frame $O_C^{\prime}$ is constant:
\begin{equation}
	{}_{O_i}^{O_C^{\prime}}\!T_m = 
	\begin{bmatrix}
		1 & 0 & 0 & 0\\
		0 & 1 & 0 & 0\\
		0 & 0 & 1 & h_{foot}\\
		0 & 0 & 0 & 1\\
	\end{bmatrix}
\end{equation}
Then the matrix of frame $O_i$ and frame $C$ can be easily obtained:
\begin{equation}
	\label{eq.17}
	{}_{O_i}^{C}\!T_m = {}_{O_C^{\prime}}^{C}\!T_m*{}_{O_i}^{O_C^{\prime}}\!T_m
\end{equation}
Now the transformation matrix of the three key frames of ESVC foot is solved analytically.

\subsection{Rollover Shape Design of Hind Foot and Fore Foot}
\label{subsec2.3}
Before deriving the hind/fore foot contact model, we need to determine the rollover shape. A common viewpoint is that segmented design is feasible for improving foot performance. Therefore, we need to answer the key questions: how to determine the number of segments, where are the segment points, and the curvature variations for each segment. Unfortunately, it is difficult to answer those questions currently. From a gait performance evaluation perspective, only by establishing a relatively accurate model of the contact points can we provide the possibility of linking the overall walking performance of the robot with the design of the rollover shape. From a design perspective, the various attributes of the robot, such as nature dynamics, target gaits, and even optimization objectives will lead to different optimal rollover shape. 

Although the profiles of rollover shape has been selected, determining the parameters remains a challenging task. Fortunately, some conclusions regarding single elliptical arc foot are provided in \cite{reference7}. Therefore, our idea is that if we can find all the parameters of ESVC foot based on partial parameters. In other words, assuming the mid ellipse is known in advance, we need a method to determine the hind/fore ellipse uniquely.

At this point, the hind ellipse still has infinitely many solutions. Without compromising design flexibility,  taking fore ellipse as the example we introduce two assumptions:

\textit{F3. At the segment point$S_2$ and $S_3$, the tangent slope of the mid ellipse should be as consistent as possible with the tangent slopes of the fore ellipses};

\textit{F4. The perpendicular bisector of $S_2S_3$ coincides with the major axis of the fore ellipses}.

We use the superscript $.^*$ to denote the situation when the contact point rolls to the boundary between the mid ellipse and the fore ellipse, i.e., the segment point $S_2$.The rollover angle of mid ellipse is denoted as $\phi_m^*$, with corresponding parameters $b_m^*$, $\phi_{mc}^*$, $l_{mrolla}^*$, $\alpha_0^*\sim\alpha_5^*$  and foot roll angle $\theta_m^*$. In $\triangle O_iO_C^{\prime}S_2$ , the distance from foot center $O_i$ to segment point $S_2$ is:
\begin{equation}
	d_0 = \left( h_{foot}^2 + {b_m^*}^2 - 2h_{foot}b_m^*\cos\alpha_0^*\right)^{\frac{1}{2}}
\end{equation}
The angle from $O_iS_2$ to the minor axis of mid ellipse is:
\begin{equation}
	\alpha_6 = \sin^{-1}\frac{b_m^*\sin\alpha_0^*}{d_0}
\end{equation}
Then the $\triangle{O_iS_2S_3}$ can be solved as:
\begin{equation}
	\left\{
	\begin{aligned}
		\alpha_7 &= \frac{\pi}{2} - \alpha_6 \\
		b_s &= \left(\frac{w_{foot}^2}{4} + d_0^2 - w_{foot}d_0\cos\alpha_7\right)^\frac{1}{2} \\
		\alpha_8 &= \sin^{-1}\frac{d_0\sin\alpha_7}{b_s} \\
	\end{aligned}
	\right.
\end{equation}
For the fore ellipse, there are four unknown parameters including major and minor axis $r_{fa}$ and $r_{fb}$, ellipse rollover angle $\phi_f^*$ at segment point and the distance of $O_fS_2$, namely $d_f^*$. From assumption $F4$, the distance of $S_2S_3$ satisfies:
\begin{equation}
	\label{eq.21}
	\frac{b_s}{2} = \sin\left(\frac{\pi}{2}-\phi_f^*\right)d_f^*
\end{equation}
where $d_f^*$ can be obtain using Eq.3\label{eq.3}. As $\alpha_10=\alpha_10$, then $\theta_f^*$ and $\theta_m^*$ satisfies:
\begin{equation}
	\label{eq.22}
	\theta_f^*-\theta_m^* = \frac{\pi}{2} - \sin^{-1}\frac{d_0\sin\alpha_7}{b_s}
\end{equation}
The relationship of rollover angle $\phi_f^*$ and roll angle $\theta_f^*$ is similar to Eq.\ref{eq.1} and Eq.\ref{eq.2}. The specific expression will be provided latter. If we enforce the condition that the slopes at segment point of the two ellipses are equal, then there are four equations for the four unknown parameters. However, due to the nonlinearity, the system of equations may not have a solution. Hence, we relax the constraint as the assumption $F3$. The slope at segment point $S_2$ of the two ellipse can be calculated as:
\begin{equation}
	\left\{
	\begin{aligned}
		\eta_m^* &= \frac{r_{mb}^2}{r_{ma}^2}\tan\phi_m^*\\
		\eta_f^* &= \frac{r_{fb}^2\tan\phi_f^* - r_{fa}^2\tan\theta_f^*}{r_{fa}^2 + r_{fb}^2\tan\theta_f^*\tan\phi_f^*}\\
	\end{aligned}
	\right.
\end{equation}
Then four parameters can be solved using a nonlinear program:
\begin{equation}
	\begin{aligned}
		&\min_{\mathbf{r_{fa},r_{fb},\phi_f^*,d_f^*,w_{foot},\theta_f^*}} \quad && w_1(\eta_m^* - \eta_f^*)^2 + w_2r_{fa}^2 + w_3r_{fb}^2 + w_4(w_{foot} - w_{foot}^*)^2\\
		&s.t. &&Eq.\ref{eq.21}, Eq.\ref{eq.22};\\
		& &&\cot\phi_f^* = \frac{r_{fb}^2}{r_{fa}^2\tan\theta_f^*};\\
		& &&d_f^* = \left(\frac{r_{fa}^2r_{fb}^2}{r_{fa}^2\cos^2\phi_f^* + r_{fb}^2\sin^2\phi_f^*}\right)^\frac{1}{2};\\
		& &&\frac{w_{foot}}{2} \ge d_0\sin\alpha_6;\\
		& &&{\eta_f^*}^2 \ge {\eta_m^*}^2;\\
		& &&e_m \ge e_f;\\
		& &&r_{fa} \ge r_{fb};\\
		& &&r_{fa},r_{fb},\phi_f^*,d_f^* \ge 0.\\
	\end{aligned}
\end{equation}
The decision variables include foot width $w_{foot}$, $\theta_f^*$ and the four unknown design parameters. $w_1$, $w_2$, $w_3$ and $w_4$ are the weight of the cost function. Boundary constraints that $\phi_f^*\in(0,\frac{\pi}{2})$, $d_f^*\in(0,0.5)$ and $w_{foot}\in(0,0.2)$ should be incorporated. By adjusting the $w_1$, the smoothness of the segment point can be controlled. It is worth noting that we do not specify the foot width in advance, and through $w_4$, and nominal foot width  $w_{foot}^*$, $w_{foot}$ can also be found. 

When the mid ellipse is known, we can uniquely determine the fore/hind ellipse using the nonlinear program. This does not mean that we lose the design flexibility. In other words, as we do not enforce the location of segment point $S_2$(equivalent to $\phi_m^*$), the range of hind/mid/fore foot can be adjusted arbitrarily. Therefore, the ESVC foot can mimic any biological foot and the deficiency of single ellipse foot is compensated.

\subsection{Model Analysis of Fore Foot Hind Foot}
\label{subsec2.4}
Now, we derived the contact model of fore/hind foot, using the fore ellipse as the example. Theoretically, the roll angle of ESVC foot $\theta$ belongs $(0,\frac{\pi}{2})$ and according to Eq.\ref{eq.22}, $\theta_f$ and $\phi_f$ may large than $\frac{\pi}{2}$. Hence, the rollover angle of fore ellipse can be separated as:
\begin{equation}
	\phi_f = 
	\left\{
	\begin{aligned}
		\frac{\pi}{2} - \tan^{-1}\frac{r_{fb}^2}{r_{fa}^2\tan\theta_f},\quad &\theta_f\in\left(\theta_m^*+\frac{\pi}{2}-\alpha_{10},\frac{\pi}{2}\right)\\
		\frac{\pi}{2} + \tan^{-1}\frac{-r_{fb}^2}{r_{fa}^2\tan\theta_f},\quad&\theta_f\in\left(\frac{\pi}{2}, \pi-\alpha_{10},\right)\\
	\end{aligned}
	\right.
\end{equation}
As $\phi_f\in\left(\phi_f^*,\frac{\pi}{2}\right]$, the arc length of $S_2C$ can be calculated as:
\begin{equation}
	\label{eq.26}
	l_{froll} \approx l_{frolla} = l_{f\phi_f} - l_{f\phi_f}^*
\end{equation}
$l_{f\phi_f}$ and $l_{f\phi_f}^*$ are the rollover length of $\phi_f$ and $\phi_f^*$, respectively, and the value can be calculated as Eq.\ref{eq.7}. Then the transformation matrix of frame $O_C$ and frame $C$ of the fore ellipse is:
\begin{equation}
	{}_{O_C}^{C}\!T_f = 
	\begin{bmatrix}
		1 & 0 & 0 & 0\\
		0 & 1 & 0 & sgn(theta)(l_{mrolla}+l_{frolla})\\
		0 & 0 & 1 & 0\\
		0 & 0 & 0 & 1\\
	\end{bmatrix}
\end{equation}
To determine the transformation matrix of frame $O_i$ and frame $C$ of the fore ellipse, we need to derive the length of $O_iC$ and the angle to the horizontal axis.In $\triangle O_fS_2C$,  $\angle S_2O_fC$ and the length of $O_fC$ are:
\begin{equation}
	\left\{
	\begin{aligned}
		\alpha_9 &= \phi_f - \phi_f^*\\
		d_f &= \left( \frac{r_{fa}^2r_{fb}^2}{r_{fa}^2\cos^2\phi_f + r_{fb}^2\sin^2\phi_f} \right)^\frac{1}{2}\\
	\end{aligned}
	\right.
\end{equation}
Then the chord length of $S_2C$ and $\angle O_fS_2C$ are:
\begin{equation}
	\left\{
	\begin{aligned}
		b_f &= \left( {d_f^*}^2+d_f^2-2d_f^*d_f\cos\alpha_9 \right)^\frac{1}{2}\\
		\alpha_{11} &= \sin^{-1}\frac{d_f\sin\alpha_9}{b_f}\\
	\end{aligned}
	\right.
\end{equation}
In $\triangle O_iS_2S_3$, $\angle O_iS_2S_3$ is :
\begin{equation}
	\alpha_{12} = \pi-\alpha_8 - \alpha_7;
\end{equation}
$\angle O_iS_2O_f$ and $\angle O_iS_2C$ can be calculated by assumption $F4$:
\begin{equation}
	\left\{
	\begin{aligned}
		\alpha_{13} &= \alpha_{12} - \phi_f^*\\
		\alpha_{14} &= \alpha_{13}+\alpha_{11}\\
	\end{aligned}
	\right.
\end{equation}
Hence, the length of $O_iC$ in $\triangle O_iS_2C$ is:
\begin{equation}
	l_{fO_iC} = \left(b_f^2 + d_0^2 - 2b_fd_0\cos\alpha_{14} \right)^\frac{1}{2}\\
\end{equation}
In $\triangle O_iS_2C$, $\angle S_2O_iC$ satisfies:
\begin{equation}
	\alpha_{15} = \sin^{-1}\frac{b_f\sin\alpha_{14}}{l_{fO_iC}}\\
\end{equation}
The auxiliary angle $\alpha_{16}$ which are defined as the angle between $S2O_i$ and the horizontal axis and its complementary angle $\alpha_{17}$ can be calculated:
\begin{equation}
	\left\{
	\begin{aligned}
		\alpha_{16} &= \frac{\pi}{2} - \alpha_6 + \theta_m\\
		\alpha_{17} &= \frac{\pi}{2} + \alpha_6 - \theta_m\\
	\end{aligned}
	\right.
\end{equation}
Then,the angle between $O_iC$ and the horizontal axis is:
\begin{equation}
	\alpha_{18} = \frac{\pi}{2} - \alpha_6 + \theta_m - \alpha_{15}\\
\end{equation}
Finally, the transformation matrix of frame $O_i$ and frame $C$ when $\phi_f \in \left(\phi_f^*,\frac{\pi}{2}\right]$ is:
\begin{equation}
	\label{eq.36}
	{}_{O_i}^{C}\!T_f = 
	\begin{bmatrix}
		1 & 0 & 0 & 0\\
		0 & \cos\theta_m & \sin\theta_m & sgn(\theta)l_{fO_iC}\cos\alpha_{18}\\
		0 & -\sin\theta_m & \cos\theta_m & l_{fO_iC}\sin\alpha_{18}\\
		0 & 0 & 0 & 1\\
	\end{bmatrix}
\end{equation}
When $\phi_f \in \left( \frac{\pi}{2},\pi-\phi_f^* \right]$, the transition of the key frames can be derived similarly and will not discuss here. However, if $\theta \rightarrow \frac{\pi}{2}$, the com axis becomes parallel to the ground which means the robot will fall. Therefore, we define $\theta_f = \frac{\pi}{2}$ as the toe edge of the ESVC foot and the calculation .

\subsection{Accuracy Analysis of ESVC Foot Model}
\label{subsec2.5}
Before integrating the ESVC foot with the bipedal robot, we need to address a remaining question. In the contact model, ${}_{O_i}^{C}\!T_m$ and ${}_{O_i}^{C}\!T_f$  are the analytical form, but the rollover length calculated by Eq.\ref{eq.7} and Eq.\ref{eq.26} is an approximate result. In this section, we will analyse the error and provide a compensation.

For a more intuitive understanding, we selected three elliptical arcs with different eccentricities as the comparison subjects, varying the rollover angle from 0 to $\frac{\pi}{2}$. The major and minor axes of EA1(ESVC Foot1) (Ellipse Arc 1) are $[0.04575,0.03750]$. These values are derived from \cite{reference7}, a set of combinations with high stability and relatively small values. The major and minor axes of EA2 (ESVC Foot3) and EA3 (ESVC Foot5) are $[0.05205,0.03150]$ and $[0.06901,0.02595]$, respectively. EA3 is closer to a circle, while EA2 lies between EA3 and EA1. More importantly, the boundary parameter angle $\lambda^*$ of EA1 is equal to 1.606 which is larger than $\frac{\pi}{2}$ that the switching parameter $\xi$, determined by Eq.\ref{eq.10} is always 0. The value of $\lambda^*$ of EA 2 and EA 3 are 1.4229 and 1.2077, respectively, which indicates that the $\xi$ can be triggered during the rollover process. Nevertheless, the boundary parameter angle of EA2 is relatively large, which reduces the possibility of triggering. In contrast, the boundary parameter angle of EA1 is relatively small, making it more likely to be triggered. Therefore, EA1, EA2 and EA3 represents three typical elliptical arcs for error analysis of the ESVC foot. 

The ground truth  is are obtained by the numerical method of Eq.6. Then error percentage is used to evaluate the accuracy:
\begin{equation}
	P_{Err}=\frac{l_{rolla}-l_{roll}}{l_{roll}}\times100\%
\end{equation}

\begin{figure}[h]
	\centering
	\includegraphics[width=1.0\textwidth]{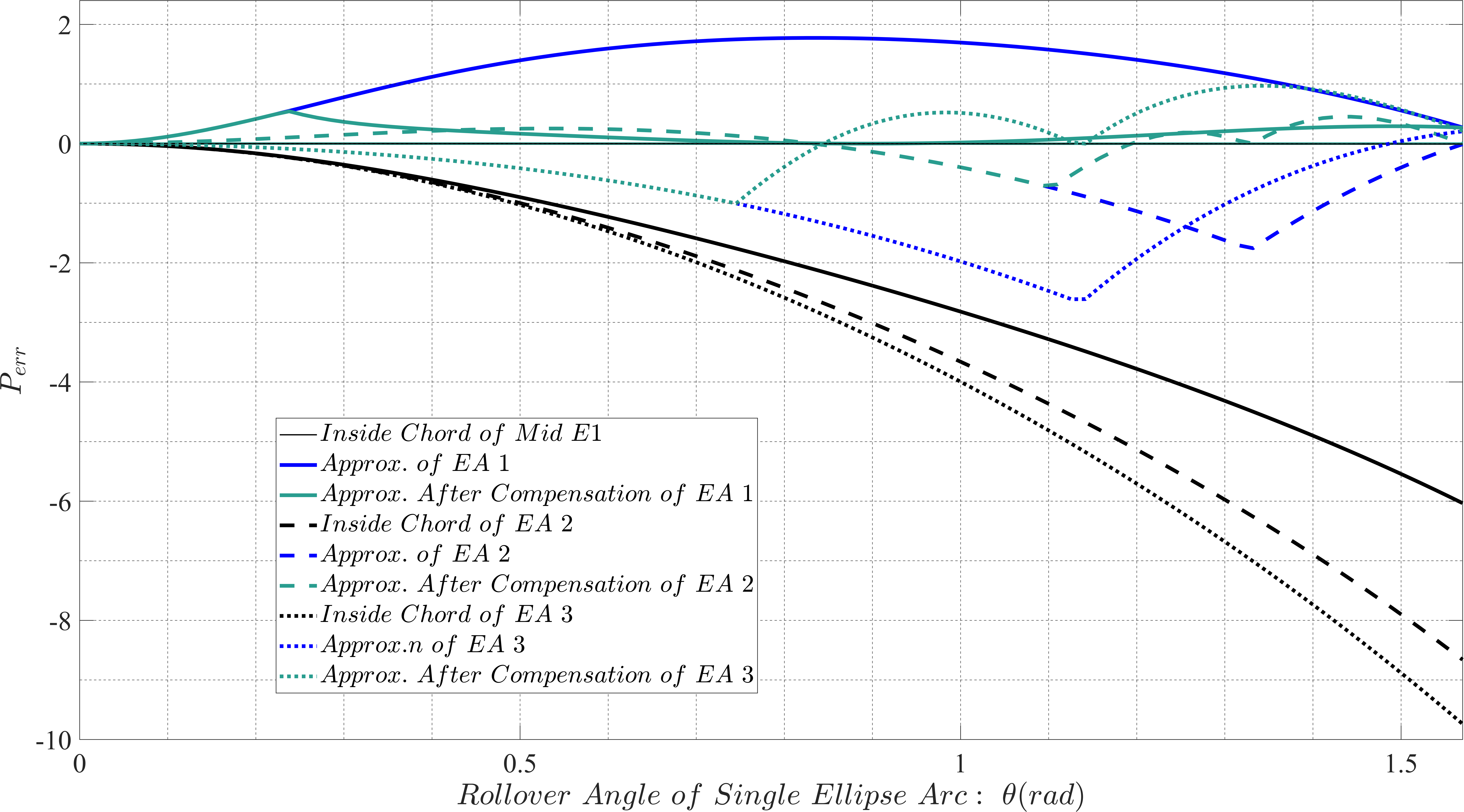}
	\caption{Accuracy Of Rollover Length For Different Ellitical Arcs}
	\label{fig3}
\end{figure}

The result of EA1, EA2, and EA3 are shown in Figure \ref{fig3}, where different line styles represent different arcs. Different color represent different approximation, that black is the inner chord approximation, blue is the approximation of Eq.\ref{eq.7}. From the comparison, we can see that both the maximum error and the average error of Eq.\ref{eq.7} are much smaller than the approximation of inside the chord. However, it is still too large for modeling contact and robot control. We also observed that the error shapes of Eq.\ref{eq.7} are all sine-like waveform. Therefore a compensation method can be proposed based on the error shape:
\begin{equation}
	\label{eq.38}
	\tilde{l}_{rolla} = 
	\left\{
	\begin{aligned}
		&l_{rolla} - sgn(\delta_{max})K_e\delta_{max}\sin\left(\frac{\lambda-2\lambda^*+\frac{\pi}{2}}{\pi-2\lambda^*}\right), &&\lambda^*<\frac{\pi}{2}\\
		&l_{rolla} - sgn(\delta_{max})K_e\delta_{max}\sin\left(\frac{\theta-2\theta_{\delta_{max}}+\frac{\pi}{2}}{\pi-2\theta_{\delta_{max}}}\right), &&\lambda^*\ge\frac{\pi}{2}\\
	\end{aligned}
	\right.
\end{equation}
$\delta_{max}$is the maximum error in the range of $[0,\frac{\pi}{2}]$ and $K_e$ is the gain for compensation. $\theta_{\delta_{max}}$ is the roll angle of maximum error when $\lambda^*\ge\frac{\pi}{2}$. Then the compensation result is depicited as the green lines in Fig\ref{fig3}.

We conclude the average error and max error in Table \ref{table1}, where it can be found that the accuracy improvement of the approximation of Eq.\ref{eq.38} is significant. For instance, considering the EA1 that used for the mid-foot, the average error of the compensation approximation is only 14.04\% to Eq.7 and the maximum error is 27.78\%. Compared to the inner chord approximation, Eq.38 has achieved an improvement of over ten times, both in terms of maximum error and average error. Additionally, in terms of error magnitude, the compensation approximation is very close to the ground truth, which is beneficial for the calculation of differential velocity. 

\begin{table}[htbp]
	\centering
	\caption{Accuracy Performance of Different Elliptical Arc}
	\renewcommand{\arraystretch}{1.8}
	\begin{adjustbox}{max width=1.0\textwidth}
		\label{table1}
		\begin{tabular}{|c|c|c|c|c|c|c|c|c|c|}
			\hline
			\makecell{\diagbox{}{}}
			&\makecell{\textbf{EA1}\\\textbf{Chord}}&\makecell{\textbf{EA1}\\\textbf{(Eq.\ref{eq.7})}}&\makecell{\textbf{EA1}\\\textbf{(Eq.\ref{eq.38})}}
			&\makecell{\textbf{EA1}\\\textbf{Chord}}&\makecell{\textbf{EA2}\\\textbf{(Eq.\ref{eq.7})}}&\makecell{\textbf{EA2}\\\textbf{(Eq.\ref{eq.38})}}
			&\makecell{\textbf{EA1}\\\textbf{Chord}}&\makecell{\textbf{EA3}\\\textbf{(Eq.\ref{eq.7})}}&\makecell{\textbf{EA3}\\\textbf{(Eq.\ref{eq.38})}}\\
			\hline
			\makecell{$\lambda^*$}
			&\makecell{1.6062}&\makecell{1.6062}&\makecell{1.6062}&\makecell{1.4229}&\makecell{1.4229}&\makecell{1.4229}&\makecell{1.2077}&\makecell{1.2077}&\makecell{1.2077}\\
			\hline
			\makecell{Max\ $\delta$}
			&\makecell{-6.04\%}&\makecell{+1.77\%}&\makecell{+0.49\%}&\makecell{-8.68\%}&\makecell{-1.57\%}&\makecell{-0.68\%}&\makecell{-9.73\%}&\makecell{-2.61}&\makecell{+0.97\%}\\
			\hline
			\makecell{Avr\ $\delta$}
			&\makecell{-2.25\%}&\makecell{+1.14\%}&\makecell{+0.16\%}&\makecell{-2.99\%}&\makecell{-0.28\%}&\makecell{-0.05\%}&\makecell{-3.29\%}&\makecell{-0.81}&\makecell{-0.07\%}\\
			\hline
		\end{tabular}
	\end{adjustbox}
\end{table}

\section{Integration of ESVC Foot And Model-Based Control}
\label{sec3}
In this section we will demonstrate how to integrated the ESVC foot with the model-based control. First, the kinematic model of the robot based on the ESVC foot is presented, and then we demonstrate how to apply the ESVC Foot within a model-based motion control framework.

\subsection{Kinematic Model of Biped Robot with ESVC Foot}
\label{subsec3.1}
Using Eq.\ref{eq.39} updates the approximation of rollover length $l_{rolla}$ according to Eq.\ref{eq.7} and Eq.\ref{eq.26}:
\begin{equation}
	\label{eq.39}
	\tilde{l}_{rolla} = 
	\left\{
	\begin{aligned}
		&\tilde{l}_{mrolla}, \quad &&\theta\in[0,\theta_m^*)\\ 
		&\tilde{l}_{mrolla} + \tilde{l}_{frolla}, \quad &&\theta\in[\theta_m^*,\alpha_{10})\\
		&\tilde{l}_{mrolla} + \tilde{l}_{f\frac{\pi}{2}} - \tilde{l}_{f\phi_f^\prime}, \quad &&\theta\in[\alpha_{10},\frac{\pi}{2})\\
	\end{aligned}
	\right.
\end{equation}
The transformation matrix of the foot frame $O_i$ and the contact frame $C$ of roll angle for mid foot and fore foot can be obtained as Eq.\ref{eq.17} and Eq.\ref{eq.36}. Here we collectively refer to ${}^C_{O_i}\!T_m$ and  ${}^C_{O_i}\!T_f$ as ${}^{Cr}_{O_i}\!T$. The superscript $r$ denotes the roll angle. As we assume the foot will not deform, when the pitch angle $\beta$ is not equal to zero, the new contact point is located at the edge in X direction and the transformation matrix is dented as ${}^{Cp}_{O_i}\!T$:
\begin{equation}
	\label{eq.40}
	{}^{Cp}_{O_i}\!T = 
	\begin{bmatrix}
		\cos\beta   & 		    0& \sin\beta   	& 	-\frac{l_{foot}}{2}\cos\beta\\
		0 		    & 		    1& 0			& 	0\\
		-\sin\beta  & 		    0& \cos\beta 	& 	\frac{l_{foot}}{2}\cos\beta\\
		0 			& 		    0& 0			& 	1\\
	\end{bmatrix}
\end{equation}
For the yaw angle, the transformation matrix ${}^{Cy}_{O_i}\!T$ is easy to write:
\begin{equation}
	{}^{Cp}_{O_i}\!T = 
	\begin{bmatrix}
		\cos\gamma   & -\sin\gamma& 0  	& 0\\
		\sin\gamma	 & 	\cos\gamma& 0	& 0\\
		0	 		 &	   		 0& 1 	& 0\\
		0 			 & 		     0& 0	& 1\\
	\end{bmatrix}
\end{equation}
Then the entire transformation matrix of foot frame $O_i$ and the contact frame $C$ is:
\begin{equation}
	{}^{C}_{O_i}\!T_{ESVC} = {}^{Cy}_{O_i}\!T*{}^{Cp}_{O_i}\!T*{}^{Cr}_{O_i}\!T
\end{equation}
Similarly, the transition of the fixed frame $O_C$ and the contact frame $C$ is 
\begin{equation}
	{}^{C}_{O_C}\!T_{ESVC} = {}^{Cy}_{O_C}\!T*{}^{Cp}_{O_C}\!T*{}^{Cr}_{O_C}\!T
\end{equation}
Then the position of COM and the pre-contact of swing foot respect to contact frame is obtained:
\begin{equation}
	\left\{
	\begin{aligned}
		&p_{com}^C(t) = {}^{C}_{O_i}\!T_{ESVC}
		\begin{bmatrix}
			f(q_i(t)) \\
			1 \\
		\end{bmatrix}\\ 
		&p_{sw}^C(t) = {}^{C}_{O_i}\!T_{ESVC}
		\begin{bmatrix}
			g(q_i(t),\theta_{sw}) \\
			1 \\
		\end{bmatrix}\\
	\end{aligned}
	\right.
\end{equation}
$f(\cdot)$represents the com position relative to support foot center frame which is only depending on joint position and $g(\cdot)$ calculates the position of pre-contact point. $\theta_{sw}$ is the roll angle of swing foot which is used to conduct the same calculation for the pre-contact point position respect to swing foot center frame. ${}^{C}_{O_i}\!T_{sp}$ is the transformation matrix of the foot center frame and the contact frame , with the subscript $sp$ representing support foot.
For the coordinates with respect to the absolute frame, it is also easy to obtain through the transformation matrices of the key frame:
\begin{equation}
	\left\{
	\begin{aligned}
		&p_{com}^{O_C}(t) = {{}^{C}_{O_C}\!T_{ESVC}}^{-1} * {}^{C}_{O_i}\!T_{ESVC}
		\begin{bmatrix}
			f(q_i(t)) \\
			1 \\
		\end{bmatrix}\\ 
		&p_{sw}^{O_C}(t) = {{}^{C}_{O_C}\!T_{ESVC}}^{-1} * {}^{C}_{O_i}\!T_{ESVC}
		\begin{bmatrix}
			g(q_i(t),\theta_{sw}) \\
			1 \\
		\end{bmatrix}\\
	\end{aligned}
	\right.
\end{equation}

\subsection{H-LIP Control with ESVC Foot}
\label{subsec3.2}
Hybrid Linear Inverted Pendulum(HLIP)\cite{reference30} is a model based control method which has been applied on Cassie, demonstrating excellent performance.There are two phases: single support phase(SSP) and double support phase(DSP), and for simplicity, we set the duration of DSP to zero.

HLIP is a linear model that the target step length is determined by the desired step length, the desired HLIP state and the robot pre-impact state for the next step, which is predicted by Step to Step(S2S) dynamics:
\begin{equation}
	\label{eq.46}
	\left\{
	\begin{aligned}
		&X_k^R=AX_{k-1}^R+Bu_{k-1}^R\\ 
		&u_k^R=u_k^h+K(X_k^R-X_k^h)\\ 
	\end{aligned}
	\right.
\end{equation}
$X_{k-1}^R$ is the impact state of the robot relative to the support foot location for current step and $u_{k-1}^R$ is the executed step length. $A$ and $B$ are the sate transformation matrix and input matrix of S2S Dynamics. $X_k^h$ and $u_k^h$ are the desired HLIP state and desired step length derived from stabilization of S2S dynamics, respectively. $u_k^R$ the is the target step length of the robot stabilization.

To verify ESVC foot, we require the robot's behavior with the ESVC foot to satisfy the HLIP constraints. In other words, the gait generation and feedback control carried out with respect to the contact frame. Taking single step as instance, the feedback pre-impact state of COM $X_{fdcom}^C$ can be calculated as:
\begin{equation}
	{\begin{bmatrix}
		p_{fdcom}^C & 1 & v_{fdcom}^C & 1\\
	\end{bmatrix}
	}^T
	=
	\begin{bmatrix}
		{}^{C}_{O_i}\!T_{sp} & 0\\
		\dot{{}^{C}_{O_i}\!T_{sp}} & {}^{C}_{O_i}\!T_{sp}\\
	\end{bmatrix}
	{\begin{bmatrix}
			p_{fdcom}^{O_i} & 1 & v_{fdcom}^{O_i} & 1\\
		\end{bmatrix}
	}^T
\end{equation}
$\dot{{}^{C}_{O_i}\!T_{sp}}$ is the time derivative of ${}^{C}_{O_i}\!T_{sp}$. Although the matrix is composed of elementary functions, its time derivative is too long to program. Therefore, in practical applications, we often use numerical methods to obtain it. Substituting $X_{fdcom}^C$ into Eq.\ref{eq.46}, the target step length with ESVC foot is obtained. Then the target swing foot trajectory  $X_{swC}^*(t)$ is generated by segmented Cubic Hermite Spline. The state swing foot pre-contact point $X_{fdsw}^C(t)$ can be achieved similarly. Then the swing foot of robot is stabilized by Eq.\ref{eq.48} and $K_{swa}$ is the correct gain. Joint level controller is identical that in \cite{reference30}.
\begin{equation}
	\label{eq.48}
	x_{swC}^{d}(t) = x_{swC}^{*}(t) + K_{swa}(X_{fdsw}^C(t)-x_{swC}^{*}(t))
\end{equation}

\section{Implementation and Experiment}
\label{sec4}
To validate the proposed ESVC Foot, simulations and physical experiments of three ESVC feet described in Section \ref{subsec2.5} are both conducted. Additionally, and for comparison, we also conduct the experiments with line foot and flat foot, as shown in Figure\ref{fig_TTII_ESVC_total}. 
\subsection{Simulation}
\label{subsec4.1}
The ESVC Feet are equipped onto the TT II robot and straight walking simulations are conducted for stability using CoppeliaSim 4.5.1. The robot employed a control scheme based on the H-LIP model, as described in Section\ref{subsec3.2}, to generate the desired step length. The center of mass(CoM)height during walking was maintained at 0.70 m, and the gait period was set to 0.38s. The pre-contact point of the ESVC Foot, which depends on the roll angle, was selected as the control point of the swing foot, and its target trajectory was generated using a segmented Hermite cubic spline. The desired CoM trajectory was analytically determined by the H-LIP model with respect to the support foot contact frame.
\begin{figure}[h]
	\centering
	\includegraphics[width=1.0\textwidth]{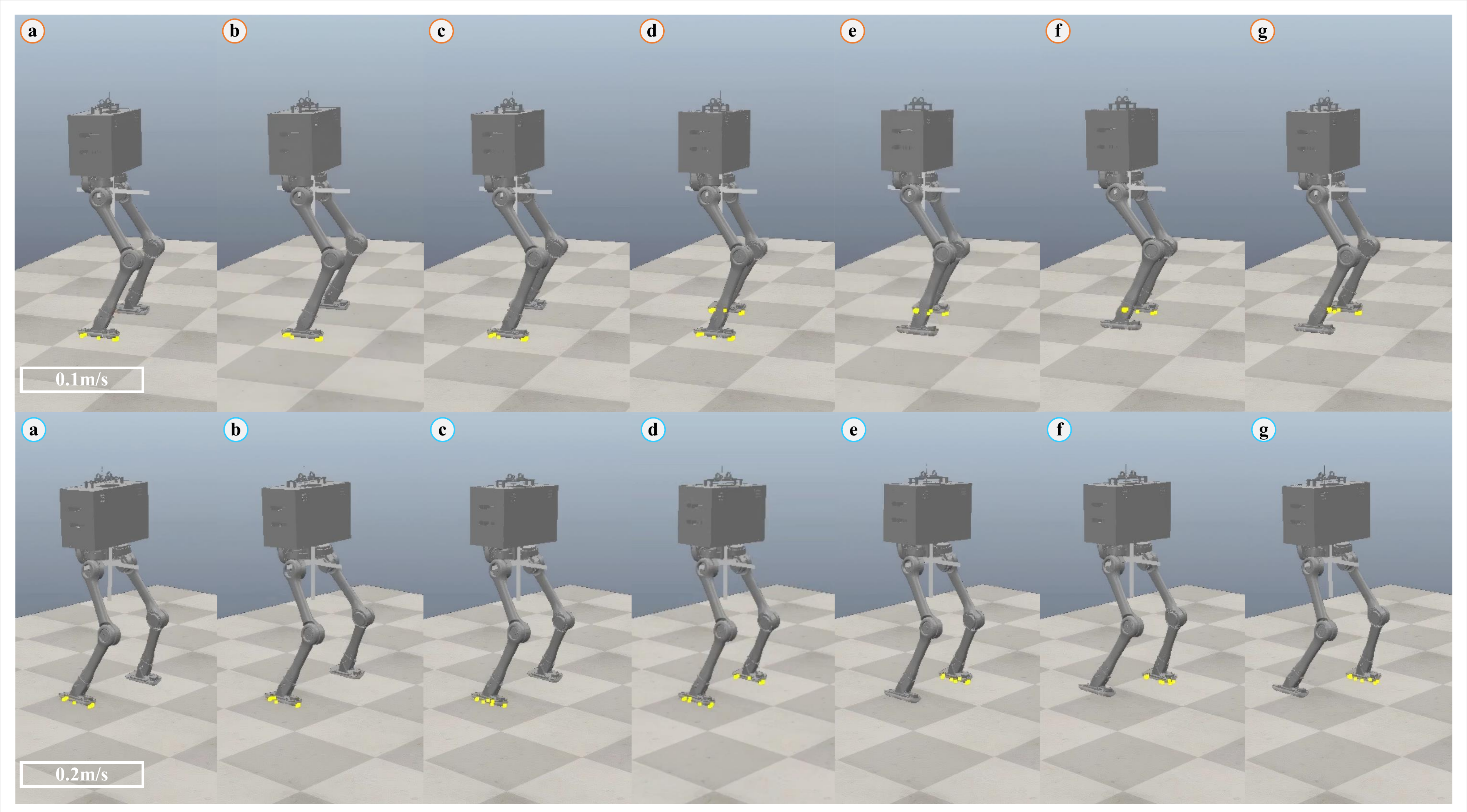}
	\caption{Snapshots of the simulation at walking speeds of 0.1m/s and 0.2m/s.}
	\label{fig4}
\end{figure}

The target joint-space trajectories were solved using the inverse kinematics(IK) solver provide by DRAKE, which is based on constrained optimization. Two separate IK solvers were designed for different support phases. The swing foot trajectory was treated as a hard position constraint, while the CoM trajectory, due to modeling errors, was not enforced as a constraint in the joint trajectory optimization. Instead, it was included as a cost term in the objective function. As discussed in Section\ref{sec2}, we decoupled the swing foot and CoM trajectories into internal and external spaces, respectively, to reduce the computational burden of the phase-wise inverse kinematics(PIK). Finally, gravity compensation and feedforward PD control were applied to track the joint trajectories.

We validated sagittal plane walking at speeds of 0.1m/s and 0.2m/s, as illustrated in Figure\ref{fig4}. The complete simulation videos are available on GitHub. These results demonstrate that the proposed ESVC Foot and modeling approach enable stable locomotion of the robot.

\subsection{Physical Evaluation}
\label{subsec4.2}
\subsubsection{Experimental Setup}
\label{subsec4.2.1}
To evaluate the impact of the ESVC Foot on the energy efficiency of robot gait, we conducted three physical experiments on the TT II robot: marking time, straight walking, and lateral walking. Each joint of the TT II robot is actuated by the high-power AK-8064 motors developed by Cubemars. The onboard main controller is an Intel i7-8665U processor running the Xenomai 3.2 real-time Linux operating system. The communication network is based on the CAN bus, and all sensor data are collected and returned by the HWT-9073-CAN device developed by WIT MOTION. Each ESVC foot is equipped with four pressure sensors for support phase detection. The HLIP-based onboard control system is implemented using a message-queue-based multi-task real-time framework. More details can be found in our previous work\cite{reference31}.
\begin{figure}[!h]
	\centering
	\includegraphics[width=0.95\textwidth]{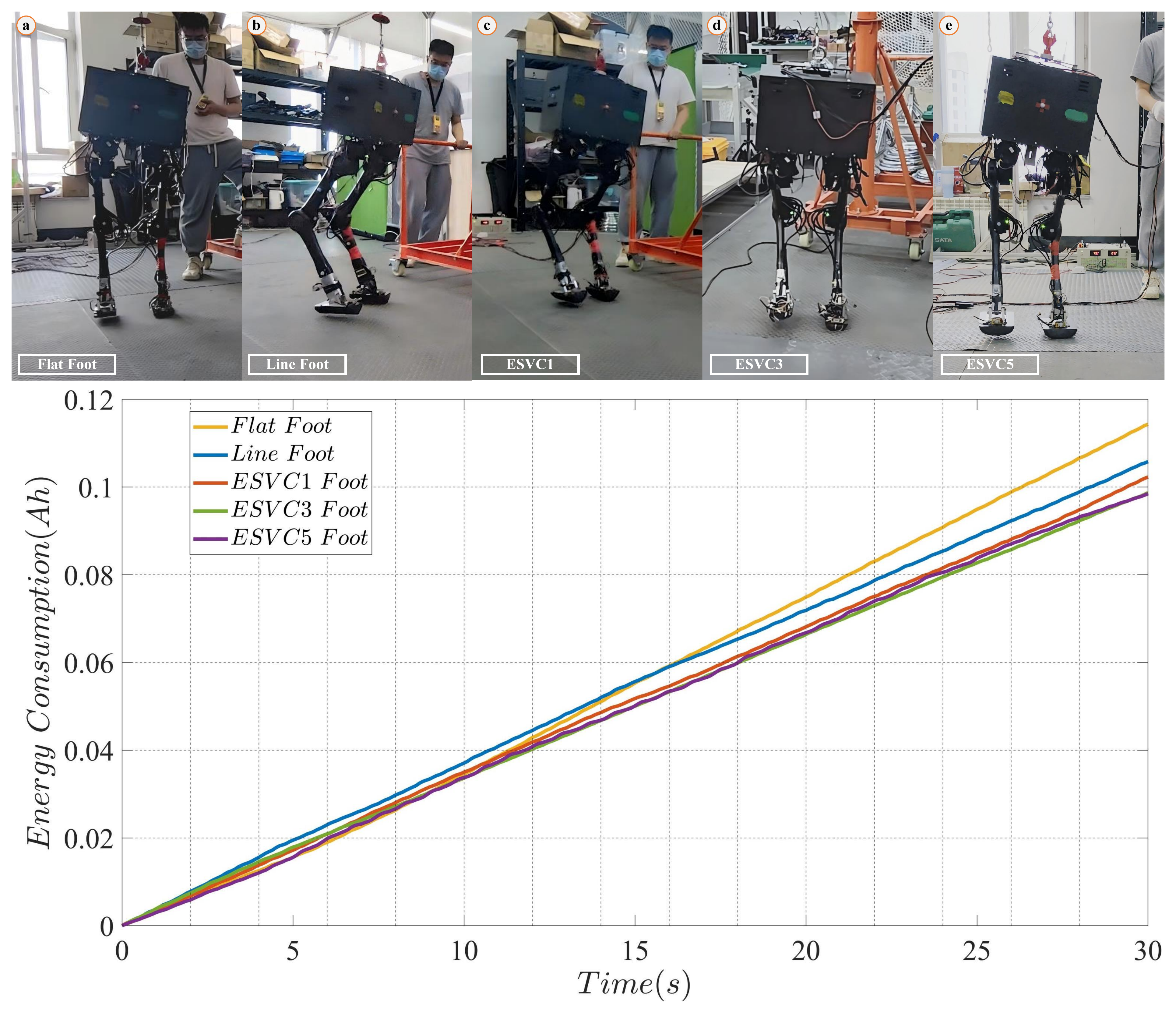}
	\caption{a.Snapshots of marking time of robot TT II; b.The Total Energy Input;}
	\label{fig5}
\end{figure}
\subsubsection{Marking Time}
\label{subsec4.2.2}
We conducted marking-time experiments on five different foot types: Flat Foot, Line Foot, ESVC1 Foot, ESVC2 Foot, and ESVC5 Foot. As shown in the upper subgraph of Figure\ref{fig5}, all five foot types enabled the robot to achieve stable marking-time walking. The lower subgraph Figure\ref{fig5} presents the energy consumption over 30 seconds of stable locomotion. It can be observed that, for marking-time walking, the Flat Foot exhibits the highest cumulative energy consumption, while the Line Foot performs better than the Flat Foot. The ESVC Foot types consume less energy compared to both the Flat and Line Feet. Among the ESVC variants, the flatter ESVC5 shows the lowest energy consumption, whereas ESVC1 has the highest.
\subsubsection{Straight Walking}
\label{subsec4.2.3}
\begin{figure}[!h]
	\centering
	\includegraphics[width=1.0\textwidth]{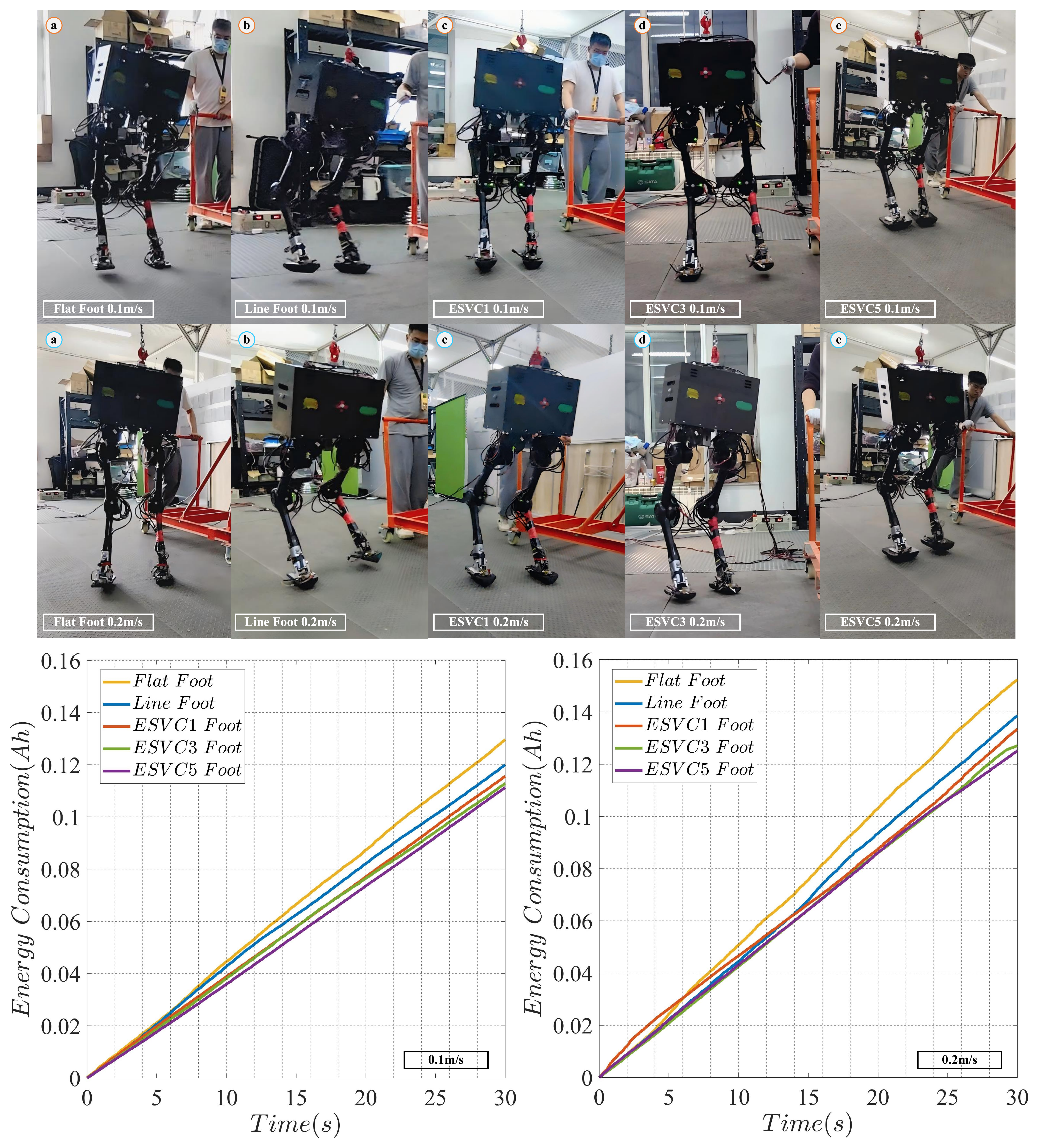}
	\caption{a.Snapshots of straight walking of robot TT II; b.The Total Energy Input Of walking speeds of 0.1m/s; c.The Total Energy Input Of walking speeds of 0.2m/s;}
	\label{fig6}
\end{figure}
We conducted sagittal plane straight walking experiments at speeds of 0.1m/s and 0.2m/s for five different foot types. Representative snapshots of the experiments are shown in the upper subgraph of Figure\ref{fig6}. The lower subgraph of Figure\ref{fig6} also presents the energy consumption over 30 seconds of walking. The energy efficiency trends are similar to those observed in the marking-time experiments: the ESVC5 Foot demonstrates the highest gait efficiency, while the Flat Foot shows the lowest.

\subsubsection{Lateral Walking}
\label{subsec4.2.4}
\begin{figure}[!h]
	\centering
	\includegraphics[width=1.0\textwidth]{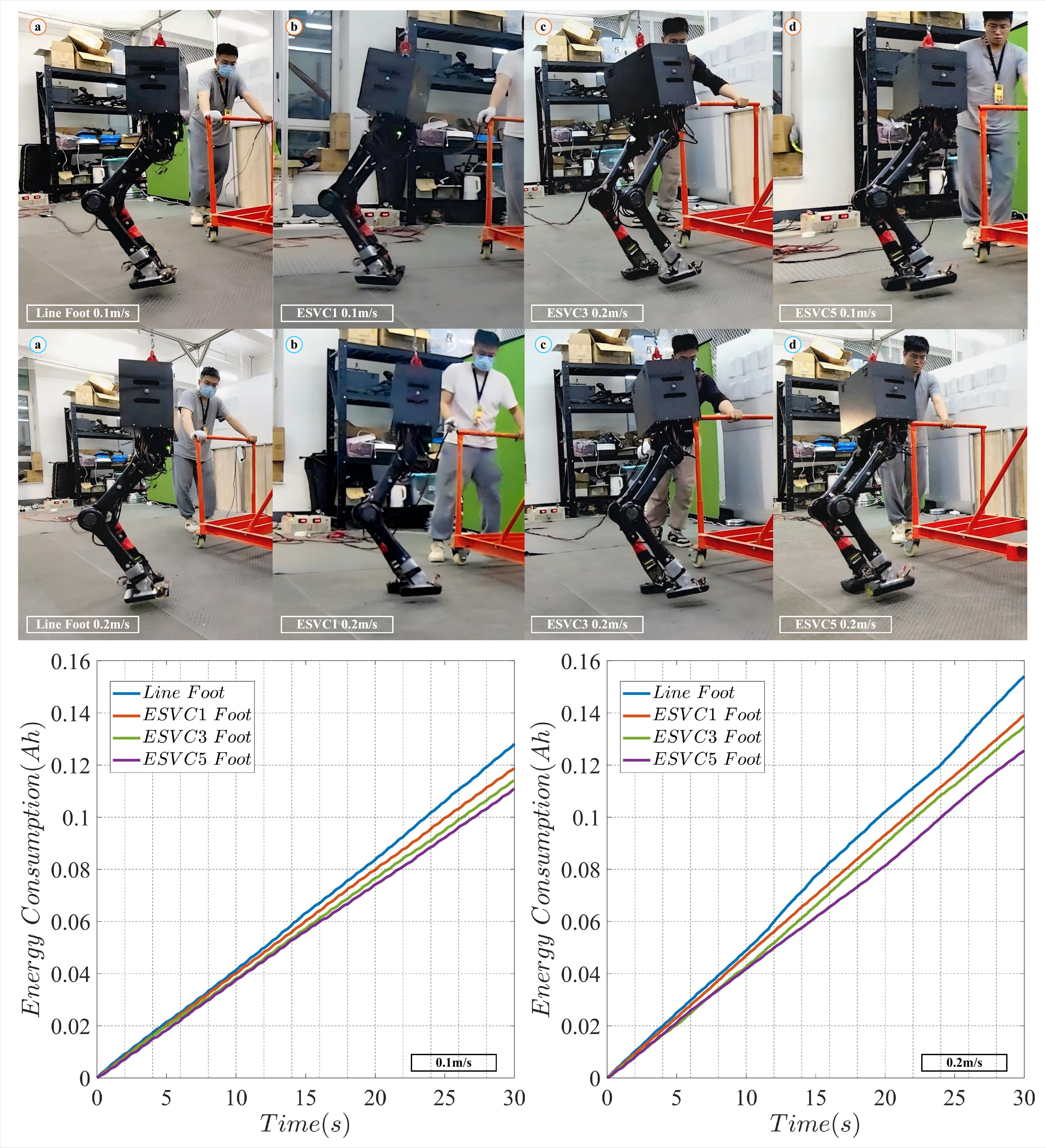}
	\caption{a.Snapshots of lateral walking of robot TT II; b.The Total Energy Input Of walking speeds of 0.1m/s; c.The Total Energy Input Of walking speeds of 0.2m/s;}
	\label{fig7}
\end{figure}
Since the primary function of the ESVC Foot lies in the frontal plane, we conducted physical experiments on lateral walking, as shown in the Figure\ref{fig7}. Due to the 5-DOF leg structure of the TT II robot, which lacks roll motion at the ankle, it is difficult to maintain frontal-plane balance when using the Flat Foot. Moreover, results from the marking-time and straight walking experiments indicate that the energy consumption of the Flat Foot is close to, and generally higher than, that of the Line Foot. Therefore, in the lateral walking experiments, we excluded the Flat Foot and selected the Line Foot as the baseline for comparison with the ESVC Feet. The lower subgraphs show the cumulative energy consumption of different foot types at walking speeds of 0.1m/s and 0.2m/s. The results reveal that all ESVC foot types consumed less energy than the Line Foot. Among them, the ESVC5 Foot showed the lowest energy consumption, and the improvement in gait energy efficiency brought by the ESVC design becomes more pronounced at 0.2m/s.

\subsubsection{Analysis}
\label{subsec4.2.5}
The cumulative energy consumption observed across the three physical walking experiments confirms that the ESVC-type feet achieve better gait energy efficiency compared to the Line Foot and Flat foot. However, this metric alone does not provide deeper insight into the underlying mechanisms. To further investigate, we perform a quantitative analysis based on average power($P_{avg}$).
\begin{equation}
	P_{avg} = \frac{E_{total}}{Time}
\end{equation}
The average gait power under different walking speeds and foot types is shown in the Table\ref{table2}.
\begin{table}[htbp]
	\centering
	\caption{Accuracy Performance of Different Elliptical Arc}
	\renewcommand{\arraystretch}{1.8}
	\begin{adjustbox}{max width=1.0\textwidth}
		\label{table2}
		\begin{tabular}{|c|c|c|c|c|c|}
			\hline
			\makecell{\diagbox{}{}}
			&\makecell{\textbf{Maring}\\\textbf{Time}}&\makecell{\textbf{Sagittal }\\\textbf{Walking(0.1m/s)}}&\makecell{\textbf{Sagittal}\\\textbf{Walking(0.2m/s)}}
			&\makecell{\textbf{Lateral}\\\textbf{Walking(0.1m/s)}}&\makecell{\textbf{Lateral}\\\textbf{Walking(0.2m/s)}}\\
			\hline
			\makecell{\textbf{Flat Foot}}
			&\makecell{658.49}&\makecell{746.43}&\makecell{877.87}&\makecell{\diagbox{}{}}&\makecell{\diagbox{}{}}\\
			\hline
			\makecell{\textbf{ESVC1 Foot}}
			&\makecell{589.21}&\makecell{665.74}&\makecell{768.87}&\makecell{685.35}&\makecell{801.45}\\
			\hline
			\makecell{\textbf{ESVC3 Foot}}
			&\makecell{568.60}&\makecell{650.08}&\makecell{731.89}&\makecell{656.85}&\makecell{776.74}\\
			\hline
			\makecell{\textbf{ESVC5 Foot}}
			&\makecell{567.24}&\makecell{640.90}&\makecell{721.25}&\makecell{639.01}&\makecell{723.19}\\
			\hline
		\end{tabular}
	\end{adjustbox}
\end{table}

From the table, we observe that during straight walking and marking-time walking, the Flat Foot exhibits the highest energy consumption approximately 8.04\%(0.0m/s), 7.99\% (0.1m/s), and 9.93\% (0.2m/s) higher than the Line Foot, respectively. Compared to the Flat Foot, the ESVC Feet offer improved energy efficiency over the Line Foot. Among them, ESVC1 is the closest in performance, with average power reductions of approximately 3.32\% (0.0m/s), 3.68\% (0.1m/s), and 8.39\% (0.2m/s), while ESVC5 is the most efficient, achieving reductions of about 6.93\% (0.0m/s), 7.27\% (0.1m/s), and 9.68\% (0.2m/s). However, the differences among the foot types are relatively small under these walking conditions, particularly between the ESVC Feet and the Line Foot. This is because, during marking-time and straight walking, the robot’s motion in the frontal plane is limited.

Among the ESVC variants, ESVC5 demonstrates the highest energy efficiency, while ESVC1 performs the worst. This is due to ESVC1 having the largest curvature, resulting in the smallest roll angle range during stance, making it functionally similar to the Line Foot when center-of-mass fluctuation is small. In contrast, ESVC5, with the smallest curvature, allows the contact point to slide further along the rollover shape, which reduces impact energy dissipation. Additionally, this configuration more closely resembles the rollover geometry of biological feet, thereby achieving the highest energy efficiency.

During frontal-plane (lateral) walking, we observe more significant performance improvements with the ESVC Feet. At a lateral walking speed of 0.1m/s, compared to the Line Foot, ESVC1, ESVC3, and ESVC5 reduce average power consumption by 7.28\%, 11.01\%, and 13.32\%, respectively. At 0.2m/s, these reductions are 6.80\%, 12.49\%, and 18.52\%, respectively. In this case, ESVC3 and ESVC5 demonstrate the most substantial improvements in energy efficiency.

Overall, the ESVC Foot designs achieve lower energy consumption compared to the Line Foot and Flat Foot. Moreover, ESVC variants with a relatively flatter midfoot curvature tend to yield greater improvements in gait efficiency. In other words, the rollover shape of these ESVC Feet more closely approximates that of a biological foot, resulting in superior energy performance during walking.

\section{Conclusion And Future Work}
This paper presents a comprehensive study on the modeling and design of a segmented varying curvature (ESVC) elliptical-arc foot for bipedal robots. First, we analyzed the spatial kinematic model of the contact point during the rollover motion of a single elliptical arc foot and derived homogeneous transformation matrices between key coordinate frames based on elementary functions. On this basis, we proposed the design of an Ellipse-based Segmented Varying Curvature Foot (ESVC Foot). Second, we investigated the modeling errors associated with the ESVC Foot and developed a compensation method. Finally, we demonstrated how the ESVC Foot can be integrated with a model-based locomotion control framework and evaluated its effectiveness in improving gait energy efficiency through physical experiments. The proposed ESVC Foot significantly enhances gait energy efficiency and extends the operational duration of bipedal robots, which is crucial for their deployment in practical applications. Moreover, the modeling approach for the ESVC Foot provides sufficient accuracy to enable stable walking and introduces a design principle that breaks the coupling between the major and minor axes of a single elliptical arc foot, laying the groundwork for optimizing foot geometry across various footed robot platforms. While this work provides meaningful insights into the modeling, design, and application of ESVC Feet, it also raises several open questions. First, although we propose a novel ESVC modeling and construction strategy based on elliptical geometry, we did not perform comparisons with alternative methods. This is partly due to the limited existing research on ESVC Feet, but more importantly, because of the lack of quantitative evaluation metrics. For instance, even when aiming to identify the energy-optimal elliptical rollover shape, we are restricted to a brute-force search for suitable major and minor axes. Thus, how to quantitatively characterize foot geometry and define optimization objectives remains a key challenge for the broader adoption of ESVC Feet. Second, this study focuses on experimental validation of the effectiveness and stability of integrating the ESVC Foot with model-based control but does not provide a theoretical derivation of simplified models under floating support conditions. We plan to address this aspect in future research. Third, the impact of ESVC Foot rollover shape and foot modeling errors on system-level noise has not yet been studied. Future work will explore noise analysis and the development of appropriate state estimators for ESVC Foot-equipped robots.

\appendix
\section{Example Appendix Section}
\label{app1}

Appendix text.






\end{document}